\documentclass[journal]{IEEEtran}
\usepackage{tikz}
\usepackage{neuralnetwork}
\usepackage{placeins}

\usepackage[utf8]{inputenc} 
\usepackage[T1]{fontenc}    
\usepackage{hyperref}       
\usepackage{url}            
\usepackage{booktabs}       
\usepackage[tableposition=top]{caption}
\usepackage{amsfonts}       
\usepackage{nicefrac}       
\usepackage{amsmath}
\usepackage{microtype}      
\usepackage{lipsum}
\usepackage[english]{babel}
\usepackage[utf8]{inputenc}
\usepackage{amsmath}
\usepackage{amsfonts}
\usepackage{graphicx}
\usepackage[colorinlistoftodos]{todonotes}
\usepackage{algorithm}
\usepackage{algpseudocode}
\usepackage{multirow}
\usepackage{graphicx}
\usepackage{booktabs}

\usepackage{makecell} 
\usepackage{authblk}
\setcellgapes{7pt}

\begin{document}

\title{Neural Network Based Undersampling Techniques}

\author{Md.~Adnan~Arefeen,
        Sumaiya~Tabassum~Nimi,
        and~M~Sohel~Rahman
\thanks{Md. Adnan Arefeen and Sumaiya Tabassum are with the Department
of Computer Science and Engineering, United International University, Bangladesh, e-mail: (adnanarefeen@gmail.com,sumtamnimi@gmail.com}
\thanks{M Sohel Rahman  is with the Department
of Computer Science and Engineering, Bangladesh University of Engineering and Technology, e-mail: sohel.kcl@gmail.com}

}

\maketitle

\begin{abstract}
Class imbalance problem is commonly faced while developing machine learning models for real-life issues. Due to this problem, the fitted model tends to be biased towards the majority class data, which leads to lower precision, recall, AUC, F1, G-mean score. Several researches have been done to tackle this problem, most of which employed resampling, i.e. oversampling and undersampling techniques to bring the required balance in the data. In this paper, we propose neural network based algorithms for undersampling. Then we resampled several class imbalanced data using our algorithms and also some other popular resampling techniques. Afterwards we classified these undersampled data using some common classifier. We found out that our resampling approaches outperform most other resampling techniques in terms of both AUC, F1 and G-mean score.
\end{abstract}


%
\IEEEpeerreviewmaketitle

\section{Introduction}

\IEEEPARstart{M}{any} real-life problems like diagnosis of diseases~\cite{breastcancer}, weather prediction~\cite{weather}, fraud detection~\cite{Frauddetect} etc. can be modelled as classification problems and can be tackled by developing machine learning models. However, in most cases, it is found that the data obtained are not balanced, that is it is not possible to collect the same number of samples for all the classes, thereby making the resulting data set class imbalanced. This problem of imbalance poses serious challenges towards developing machine learning models as follows. The models become biased towards the majority class and hence mostly fail to detect minority classes. Therefore, in such a scenario, despite obtaining a good accuracy, we do not obtain good scores in terms of other metrics of performance like F1 score~\cite{fmeasure}, AUC score~\cite{AUC}, G-mean score~\cite{g-mean} etc. Since this issue of class imbalance is challenging and damaging, significant attention has been given to solve this issue in the literature~\cite{imb-review}. The most common methods among these use resampling techniques to bring balance in the dataset. Resampling can be done by reducing the number of majority class samples. This technique is popularly known as undersampling. Some common undersampling techniques include cluster centroids~\cite{imblearn}, tomek's links~\cite{tomek}, neighbourhood cleaning rule~\cite{ncl} etc. Resampling can also be done by increasing the number of minority class samples by either duplicating some data or generating new data. This technique is called oversampling. SMOTE~\cite{smote} and several variants of SMOTE~\cite{smote-1}~\cite{smote-2}, ADASYN~\cite{adasyn} etc. are some frequently used oversampling techniques. \newline
In spite of significant works in this area in the literature, there are little scope for much improvement. With this backdrop, we revisit neural network based approaches for under-sampling. Neural networks have been successfully used for tasks like image recognition, natural language processing etc. in recent years. We explored the possibility of using the potentials of neural networks to capture intricate patterns within data to solve the issue of class imbalance.

\section{Related Work}
Class imbalance, being a challenging problem, has attracted many researcher’s attention throughout the past and recent years. To bring balance in the imbalanced data, strategies like oversampling and undersampling of data were employed. These researches were conducted as early as 1972. A popular algorithm for undersampling, Edited Nearest Neighbor (ENN) rule was proposed in the paper \cite{enn}. ENN works by removing the data points whose class label does not match the majority of its k nearest neighbors. Another popular algorithm for undersampling, Tomek links removal (TLL) was introduced in \cite{tomek}. This algorithm works by detecting pair of data points, called Tomek link, that are each other’s nearest neighbor but have different class labels. Undersampling can be done by either removing all Tomek links or by
removing the majority class data belonging to the Tomek link. The NearMiss (NM) methods perform undersampling by removing data points from majority class based on their distances between each other \cite{nearmiss}. In NearMiss-1, the points in majority class whose mean distance to the k-nearest points in minority class is lowest are retained, where k is a tunable hyperparameter. Whereas, NearMiss-2 retains those points from majority class whose mean distance with k farthest points in minority class is lowest. In the final version of NearMiss, NM-3, for every data point in minority class, k nearest data points in minority class are retained. In addition to these undersampling techniques, there is another undersampler called clustered centroids \cite{imblearn} which makes use of k-means clustering to balance an imbalanced dataset by reducing the number of majority samples.   

\section{Methods}
\begin{figure*}[!htbp]
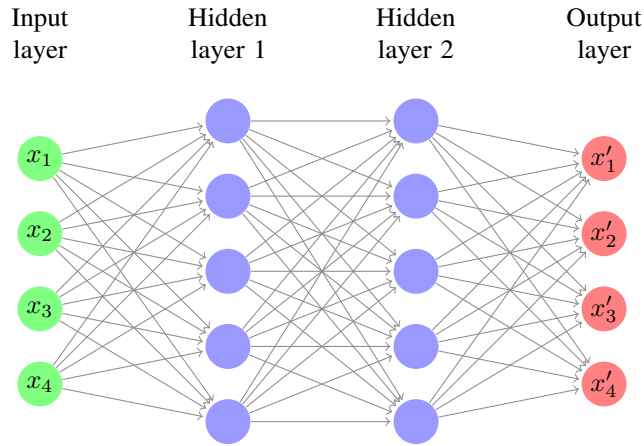

  \centering

\begin{neuralnetwork}[height=5.5]
		\newcommand{\nodetextclear}[2]{}
		\newcommand{\nodetextx}[2]{$x_#2$}
		\newcommand{\nodetexty}[2]{$x'_#2$}
		\inputlayer[count=4, bias=false, title=Input\\layer, text=\nodetextx]
		\hiddenlayer[count=5, bias=false, title=Hidden\\layer 1, text=\nodetextclear] \linklayers
		\hiddenlayer[count=5, bias=false, title=Hidden\\layer 2, text=\nodetextclear] \linklayers
		\outputlayer[count=4, title=Output\\layer, text=\nodetexty] \linklayers
\end{neuralnetwork}
\caption{Simple Neural Network to generate input. The nodes shown in green color are inputs. Two hidden layers are shown in blue color. The output layer is shown in red color. The line between each node represents connection between each layer. The network is fully connected. There are five neurons aka nodes for each hidden layer shown in the figure.}
 \label{fig1}
\end{figure*}

\begin{figure*}[!htbp]
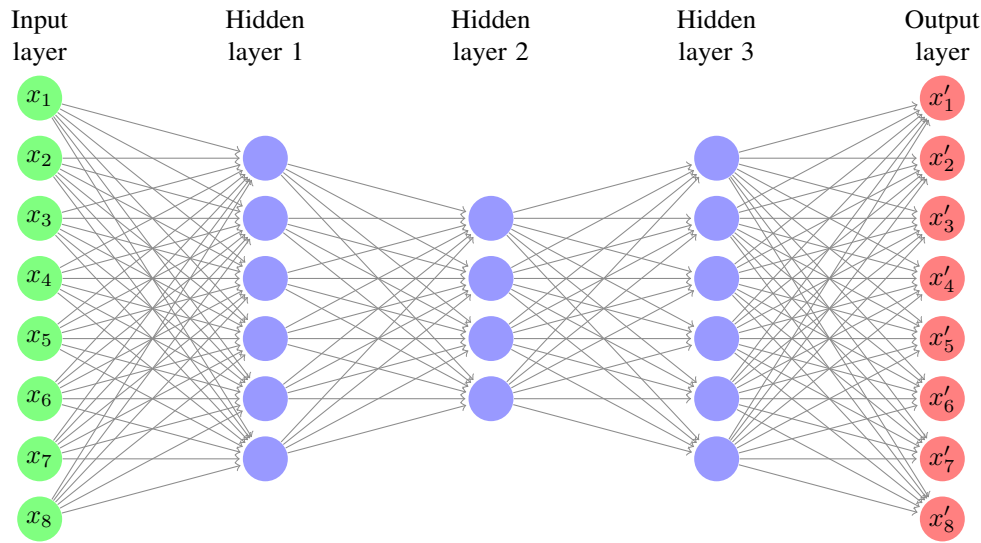

  \centering

\begin{neuralnetwork}[height=10, layertitleheight=0, nodespacing=0.8cm, layerspacing=3cm]
		\newcommand{\nodetextclear}[2]{}
		\newcommand{\nodetextx}[2]{$x_#2$}
		\newcommand{\nodetexty}[2]{$x'_#2$}
		\inputlayer[count=8, bias=false, title=Input\\layer, text=\nodetextx]
		\hiddenlayer[count=6, bias=false, title=Hidden\\layer 1, text=\nodetextclear] \linklayers
		\hiddenlayer[count=4, bias=false, title=Hidden\\layer 2, text=\nodetextclear] \linklayers
		\hiddenlayer[count=6, bias=false, title=Hidden\\layer 3, text=\nodetextclear] \linklayers
		\outputlayer[count=8, title=Output\\layer, text=\nodetexty] \linklayers
\end{neuralnetwork}

 \caption{An autoencoder neural network model. The inputs are regenerated by first encoding and then by decoding the encoded representation. The output of the middle layer is the encoded representation of the inputs. The models trains the inputs through an unsupervised approach.}
\label{fig2}

\end{figure*}

We use an auto-encoder and a simple artificial neural network for training the minority class. Figure \ref{fig1} and Figure \ref{fig2} depict two such models. We fitted the minority data using one of the two models. A threshold value was set to choose the kind of neural network to be used to train the minority samples. We have set the threshold value to 30. If the number of input attributes are more than 30, we have fitted the minority samples using an autoencoder; otherwise, we fitted those with a simple neural networks with 2/3 hidden layers. Notably, solving the issue of over-fitting was not a major concern for our task. The reason behind this is, we have to generate a minority sample with approximately 100\% accuracy. If we can not fit the minority samples well, we may loose information on predicting majority samples. If the model is not strong enough, it may propagate error when predicting majority samples.

\subsection{Undersampling Algorithm}

\subsubsection{Algorithm 1: Hard Neural Network Based Undersampling}
Suppose, we have $n_1$ minority samples and $n_2$ majority samples in the dataset under consideration. In this algorithm, we train a neural network (autoencoder or feedforward, decided based on the value of a predefined threshold as discussed in the previous section) to learn the values of features of the $n_1$ minority samples and then we use the same neural network to predict the features of the $n_2$ majority samples. Then we calculate the euclidean distance between the predicted and the real values of the features. In a list, we store the values of these euclidean distances mapped by the indices of the corresponding majority class samples. We then sort the list in descending order based on the values of the euclidean distances calculated. From this sorted list we choose first $n_1$ data samples. The final dataset obtained is the combination of $n_1$ minority class data from the original dataset and $n_1$ majority class data chosen by our approach. So in effect we choose those samples from the majority class that are far in terms of euclidean distance from the predicted values. In other words, our under-sampling approach actually removes the majority class samples which are present in the vicinity of the minority class samples and retains the majority class samples which are located further from the minority class samples. Hence the decision boundary becomes more defined and the resulting balanced dataset becomes more separable. As a consequence, this algorithm outperforms most other undersampling algorithms for most datasets. However, we noted that, this algorithm performs the best when there is no overlap between data points, as will be evident in section 5 when we will generate some artificial data points and observe the performance of the algorithm on those data. For overlapping data, we have proposed another algorithm in the next subsection. 

  \begin{algorithm}[!htb]
   \caption{Hard Under-sampling Using Neural Network}
    \begin{algorithmic}[1]

        \State ${n_1} \xleftarrow[]{}$ \textrm{number of samples of the minority class}
        
        \State ${n_2} \xleftarrow[]{}$ \textrm{number of samples of the majority class}
        \State ${m} \xleftarrow[]{} $ \textrm{number of attributes}
        
        \State $majSamples[1 \dots n_2] \xleftarrow[]{} $ \textrm{Samples of the Majority class}
        \State $minSamples[1 \dots n_1] \xleftarrow[]{} $ \textrm{Samples of the Minority class}

        \If{$m > threshold$}
            \State $model \xleftarrow[]{} autoencoder.train(minSamples[1 \dots n_1])$
        \Else
        
            \State $model \xleftarrow[]{} simpleANN.train(minSamples[1 \dots n_1])$
        \EndIf
        \State $distArray \xleftarrow[]{} \{\}$  
        \For{ \textrm{each} $x \in n_2$}                    
            \State $x' \xleftarrow[]{} model.predict(x)$
            \State $d \xleftarrow[]{} ||x - x'||^2_2$
            \State $index \xleftarrow[]{} x.index$
            \State $distArray \xleftarrow[]{} distArray \cup \{d,index\}$
        \EndFor
        
        \State $sortedList \xleftarrow[]{} sort(majSamples[1 \dots n_2],distArray)$ \Comment{\textrm{Sort the indices of $n_2$ samples according to descending order of distance}}
        \State $selectedIndices \xleftarrow[]{} sortedList[1..n_1]$ \Comment{\textrm{select first $n_1$ number of indices from the $sortedList$}}
        \State $X_1 = \{\}$ 
        \For {\textrm{each }$index \in selectedIndices$}
            \State $X_1 = X_1 \cup majSamples[index]$
        \EndFor
        \State $finalData = X_1 \cup minSamples[1 \dots n_1])$

\end{algorithmic}

\end{algorithm}

\subsubsection{Algorithm 2: Soft Neural Network Based Undersampling}
As discussed at the end of the previous subsection, our proposed Hard Neural Network Based Undersampling algorithm (NUS-1) does not perform well when there is overlap between data points in the dataset. To resolve this issue, we have proposed a new algorithm in this subsection called the soft neural network based undersampling. The soft neural network based undersampling (NUS-2) differs from hard neural network based undersampling in how the majority samples are selected. We choose exactly the first $n$ samples from majority class from the indices which are far from its predicted values. That is why we called the algorithm Hard Neural Network Based Undersampling. At first we predict the minority samples by the model that was fitted on the samples from the minority class. The maximum euclidean distance is calculated. Besides calculating the maximum distance, we also calculate the average distances of half of the samples which are greater in value than the other half of the samples. After that, we predict the samples of the majority class with the same model. This time we choose the samples of majority class as follows. We feed one sample to the model, generated its clone by the model and calculated euclidean distance between the two. If the distance is higher than the maximum distance or the half-average distance of the samples from the minority class, we include it in the final dataset as a sample of majority class. The soft neural network based undersampling algorithm performs better than all other undersampling algorithms for sampling overlapping data, as will be observed in section 5 when we will see the effect of different undersampling algorithms on artificially generated overlapping data.

  \begin{algorithm}[!htb]
   \caption{Soft Under-sampling Using Neural Network}
    \begin{algorithmic}[1]

               \State ${n_1} \xleftarrow[]{}$ \textrm{number of samples of the minority class}
        
        \State ${n_2} \xleftarrow[]{}$ \textrm{number of samples of the majority class}
        \State ${m} \xleftarrow[]{} $ \textrm{number of attributes}
        
        \State $majSamples[1 \dots n_2] \xleftarrow[]{} $ \textrm{Samples of the Majority class}
        \State $minSamples[1 \dots n_1] \xleftarrow[]{} $ \textrm{Samples of the Minority class}

        \If{$m > threshold$}
            \State $model \xleftarrow[]{} autoencoder.train(minSamples[1 \dots n_1])$
        \Else
        
            \State {$model \xleftarrow[]{} simpleANN.train(minSamples[1 \dots n_1])$}
        \EndIf
        \State $distArray \xleftarrow[]{} \{\}$  
        \For{ \textrm{each} $x \in n_2$}                    
            \State $x' \xleftarrow[]{} model.predict(x)$
            \State $d \xleftarrow[]{} ||x - x'||^2_2$
            \State $index \xleftarrow[]{} x.index$
            \State $distArray \xleftarrow[]{} distArray \cup \{d,index\}$
        \EndFor
        
        \State $sortedMinorityIndexList = $ \textrm{Sort the indices of $n_1$ samples according to descending order of distance}
        
        \State $lastMid_{avg} = $ \textrm{Average distance of half of the minority samples whose indices are in first half of $sortedMinorityIndexList$}
        \State $Maxdist= $ \textrm{Maximum distance calculated among the minority samples' with their prediction}
        \For{each sample in $n_2$}                    
            \State \textrm{Predict the majority sample's attribute using the trained model}
            
            \State \textrm{Calculate euclidean distance between real and predicted attributes of the sample}
            \State \textrm{Map this distance with index of the sample}
        \EndFor
        
        \State $sortedMajorityList = $ \textrm{Sort the indices of $n_2$ samples according to descending order of distance}
        \State $selectedIndices = [   ]$
        \For{each index in $sortedMajorityList$}
            \State $dist = MajoritySamples[index].dist$ 
            \If{$dist > Maxdist$ \textbf{or} $ dist > lastMid_{avg}$}
                \State $selectedIndices.append(index)$
            \EndIf
        \EndFor
        \State $X_1 = []$ 
        \For {each index in selectedIndices}
            \State Append($X_1,MajoritySamples[index]$)
        \EndFor
        \State $finalData = Combine(X_1,MinoritySamples[1 \dots n_1])$

\end{algorithmic}
\end{algorithm}

\section{Results Analysis}
\subsection{Overview of the experiments}
We have designed our experiments as follows. We under-sample the dataset under consideration using different undersampling algorithms. Subsequently, the under-sampled dataset is fed to a number of classifiers and we evaluate the classification results thereof. In Table~\ref{table:sampling-algo}, we list the classifiers and the undersampling algorithms we used.

\begin{table*}[!htb]
\centering
\begin{tabular}{||c||c||} 
 \hline
Undersampling Algorithms & Classfier Algorithms\\ [0.5ex] 
 \hline\hline
 Edited Nearest Neighbour (ENN)~\cite{enn} & Random forest (RF)~\cite{randomforest}\\ 
 All KNN  (AKNN)~\cite{tomek} &
 Gradient-boosting (GradBoost)~\cite{gradboost}\\
 Near Miss (NM-1 NM-2 NM-3)~\cite{nearmiss}
 & K-nearest neighbour ~\cite{knn-1}\\
 Neighbourhood Cleaning Rule  (NCR)~\cite{ncl}& Stochastic gradient descent (SGD)~\cite{sgd}\\
 Random Undersampling (RUS)
&  Logistic Regresson (LR) ~\cite{lr}\\
 Tomek Link (TLL)~\cite{tomek} & 
 \\
 \hline
\end{tabular}
\caption{Under-sampling Algorithms}
\label{table:sampling-algo}
\end{table*}

The result analysis section is organised as follows. First, we gave a little description of the dataset used in this paper. Then, we demonstrated the metrics used in the experiment for comparison. After that, we showed the results generated by various classifiers such as Gradient Boosting Classifier (GradBoost), Stochastic Gradient Descent Classifier (SGD), K-nearest neighbour classifier (KNN), Random Forest(RF) and Logistic Regression (LR). We have used scikit-learn, scipy, numpy, pandas packages to implement all these algorithms and for data conversion~\cite{scikit-learn,numpy,matplotlib,imblearn}. We have used keras package to implement the neural network and the auencoder~\cite{keras}. For graphical representation, we have used matplotlib package~\cite{matplotlib}. We made the dataset under sampled by different undersampling algorithms such as Edited Nearest Neighbour(ENN)~\cite{enn}, ALL KNN, Near Miss algorithm (Version- 1, 2 \& 3)~\cite{nearmiss}, Tomek link Undersampler (TLL)~\cite{tomek}, Random Undersampler (RUS) and the proposed 2 algorithms Neural Network Based Undersampling 1 \& 2 (NUS-1 \& NUS-2) also called hard undersampling and soft undersampling algorithms using neural network. Later, these undersampled data with binary class were classified by the classifiers stated above. We showed the metric value produced by each classifier for comparison.

\subsection{Evaluation Criteria}

For evaluating the performance of our proposed algorithm, we use some ROC (Receiver Operating Characteristics) curve~\cite{au-roc} based performance metrics. Let {+,-} represent positive and negative class labels. Table \ref{tab:conf_matrix} called confusion matrix represents performance of classification algorithm. Based on the confusion matrix in Table \ref{tab:conf_matrix} the performance metrics as defined in this section are used to evaluate learning of imbalanced data sets by our proposed algorithms.
\FloatBarrier
\begin{table}[!htb]
\makegapedcells
\caption{Confusion Matrix}
    \centering
    \begin{tabular}{cc|cc}
    \multicolumn{2}{c}{}
    &   \multicolumn{2}{c}{Predicted} \\
    &       &   + &   -              \\ 
    \cline{2-4}
\multirow{2}{*}{\rotatebox[origin=c]{90}{Actual}}
    & +   & True Positive (TP)   & False Negative (FN)                 \\
    & -    & False Positive (FP)    & True Negative (TN)                \\ 
    \cline{2-4}
\end{tabular}
    \label{tab:conf_matrix}
\end{table}

For comparing the performance of different undersampling algorithms on classification, we use the metric Area under the Receiver Operating Characteristics (ROC) curve\cite{au-roc}, the area under ROC curve is popularly known as AUC. AUC value measures the degree of separability between classes. Higher value of AUC indicates that the model is more capable of distinguishing the classes than a model with lower AUC value. The problem with imbalanced dataset is that any machine learning algorithm trained on these data becomes more biased towards the majority class. In addition, overlapping of samples from different classes also poses a problem to the performance of the model because it can not distinguish between classes. This phenomenon is reflected in lower AUC value during evaluation. Under-sampling potentially can solve the problem of imbalance by removing some samples from the majority class and thus by making the dataset more balanced. AUC value becomes higher when trained with these balanced data. In Table ~\ref{tab:Balance-auc}, ~\ref{tab:Pima-auc} and ~\ref{tab:satimage-auc} and ~\ref{tab:Ionosphere-auc}, we showed the AUC values of different machine learning models on some originally imbalanced datasets~\cite{dataset} resampled by several under-sampling techniques. The G-mean is defined as the square root of the product of true positives (TP) and false positives (FP). The equation is as follows.
\begin{equation}
    G-mean = \sqrt{TP \times TN}
    \label{eq:gmean}
\end{equation}

The F1 measure is another popular performance metric to evaluate the performance of classification algorithms which is defined as follows.
\begin{equation}
    F1 = \frac{2 \times precision \times recall}{precision + recall}
    \label{eq:F1}
\end{equation}
The terms precision and recall in this formula refer to the ratio of true positives (TP) and false positives (FP) respectively to the total number of samples, defined as follows: 
\begin{align*}
    precision = \frac{TP}{TP + FP}
    \\
    recall = \frac{TP}{TP+FN}
    \label{eq:prec-recall}
\end{align*}

 \subsection{Description of Dataset}
 We have used four real world datasets to do experiment on the proposed algorithms. All of them are from UCI machine learning repository~\cite{Dua:2019}. The imbalanced ratio is defined as $N_{maj}/N_{min}$. The description of the data sets are available in Table~\ref{table:descData}.
 \FloatBarrier
 \begin{table}[!htb]
 \caption{Description of Dataset}
\centering
\begin{tabular}{||c c c c c||} 
 \hline
Dataset & \#attribute &\#min & \#maj & Ratio \\ [0.5ex] 
 \hline\hline
 Ionosphere& 34  & 126 & 225 & 1.78 \\ 
 Balance & 4 & 49 & 576 & 11.8 \\
 Pima & 8 & 268 & 500 & 1.9 \\
 Satimage & 36 & 626 & 5809 & 9.27 \\
 \hline
\end{tabular}

\label{table:descData}
\end{table}
\FloatBarrier

The number of majority samples selected by each under sampling algorithms are described in Table~\ref{tab:MajorityDataTable}.
\FloatBarrier
\begin{table}[!htb]
\caption{Number of Majority Samples Chosen by each undersampler}
\centering
\resizebox{\linewidth}{!}{
\begin{tabular}{||c c c c c c c c c c c c||} 
 \hline
Dataset & ENN &AKNN & NM1 & NM2 & NM3 & NUS1 & NUS2 & CC & NCR & TLL & RUS\\ 
 \hline\hline
 Ionosphere& 216  & 215 & 126 & 126 & 99 & 126 & 105 &126 &146 & 225 & 126\\ 
 Balance & 452  & 427 & 49 & 49 & 49 & 49 & 161 &49 &544 & 571 & 49\\
 Pima & 279  & 249 & 268 & 268 & 268 & 268& 204 &268 &261 & 450 & 268\\
 Satimage & 5319  & 5213 & 626 & 626 & 626 & 626& 3045 &626 &5449 & 5770 & 626\\

 \hline
\end{tabular}}

\label{tab:MajorityDataTable}
\end{table}
\FloatBarrier
In almost all cases, we found that, our proposed undersamplers, NUS1 and NUS2 outperform all other undersamplers in case of almost all training algorithms. NUS1 and NUS2 resample the data in such a way that they become more separable as noted from Figures \ref{fig:fig2} and \ref{fig:fig3}. This leads to higher AUC, G-mean and F1 values and hence better performance. It is to be noted that we have used a number of classifiers to verify that the proposed undersampling algorithms are not classifier dependent. 
\FloatBarrier
 
\begin{table}[!htb]
 \caption{AUC values of Balance dataset using various classifiers}
  \centering
  \resizebox{\linewidth}{!}{
  \begin{tabular}{llllll}
    \toprule
    \multicolumn{6}{c}{Balance Dataset}                   \\
    \midrule
    Method     & GradBoost     & SGD & KNN & RF & LR\\
    \midrule
    ENN & $0.498 \pm 0.060$  
    & $0.500 \pm 0.006$  & 
    $0.554 \pm 0.111$ &   
    $0.498 \pm 0.030$ & 
    $0.500 \pm 0.000$\\
    AKNN     & $0.511 \pm 0.063$ & 
    $0.500 \pm 0.000$  & 
    $0.595 \pm 0.090$ &  
    $0.516 \pm 0.058$ & 
    $0.500 \pm 0.000$  \\
    NM1     & $0.417 \pm 0.193$    & 
    $0.497 \pm 0.145$ & 
    $0.513 \pm 0.188$ & 
    $0.278 \pm 0.147$ & 
    $0.526 \pm 0.179$\\
    NM2     & $0.783 \pm 0.155$    & 
    $0.496 \pm 0.155$ &
    $0.744 \pm 0.144$ &  
    $0.772 \pm 0.157$ &
    $0.415 \pm 0.190$\\
    NM3     & $0.447 \pm 0.152$   & 
    $0.494 \pm 0.129$ & 
    $0.636 \pm 0.202$ & 
    $0.366 \pm 0.209$ & 
    $0.492 \pm 0.175$\\
    NUS1     & $0.887 \pm 0.119$   & 
    $\textbf{0.860} \pm 0.215$ & 
    $\textbf{0.971} \pm 0.086$ &
    $0.897 \pm 0.117$ &
    $\textbf{0.897} \pm 0.122$\\
    NUS2     & $0.809 \pm 0.148$   & 
    $\textbf{0.768} \pm 0.181$ & 
    $\textbf{0.842} \pm 0.114$ & 
    $0.796 \pm 0.128$ & 
    $\textbf{0.721} \pm 0.143$\\
    CC & $\textbf{0.964} \pm 0.087$   & 
    $0.476 \pm 0.142$ & 
    $0.582 \pm 0.162$ & 
    $\textbf{0.944} \pm 0.098$ & 
    $0.359 \pm 0.154$\\
    NCR     & $0.494 \pm 0.022$   & 
    $0.501 \pm 0.008$ & 
    $0.518 \pm 0.066$ & 
    $0.497 \pm 0.008$ & 
    $0.500 \pm 0.000$\\
    TLL     & $0.495 \pm 0.013$   & 
    $0.500 \pm 0.000$ & 
    $0.496 \pm 0.031$ & 
    $0.493 \pm 0.010$ & 
    $0.500 \pm 0.000$\\
    RUS     & $0.561 \pm 0.181$   & 
    $0.519 \pm 0.099$ & 
    $0.657 \pm 0.201$ & 
    $0.522 \pm 0.201$ & 
    $0.487 \pm 0.194$\\
    \bottomrule
  \end{tabular}}
  \label{tab:Balance-auc}
\end{table}

\FloatBarrier
\begin{table}[!htb]
 \caption{G-Mean values of Balance dataset using various classifiers}
  \centering
  \resizebox{\linewidth}{!}{
  \begin{tabular}{llllll}
    \toprule
    \multicolumn{6}{c}{Balance Dataset}                   \\
    \midrule
    Method     & GradBoost     & SGD & KNN & RF & LR\\
    \midrule
    ENN & $0.081 \pm 0.275$  & 
    $0.000 \pm 0.000$  & 
    $0.313 \pm 0.366$ &   
    $0.019 \pm 0.151$ & 
    $0.000 \pm 0.000$\\
    AKNN     & $0.211 \pm 0.340$ & 
    $0.000 \pm 0.000$  & 
    $0.451 \pm 0.276$ &  
    $0.147 \pm 0.340$ & 
    $0.000 \pm 0.000$  \\
    NM1     & $0.366 \pm 0.185$    & 
    $0.192 \pm 0.471$ & 
    $0.483 \pm 0.211$ & 
    $0.213 \pm 0.230$ & 
    $0.513 \pm 0.209$\\
    NM2     & $0.766 \pm 0.178$    & 
    $0.103 \pm 0.360$ &
    $0.724 \pm 0.158$ &  
    $0.780 \pm 0.161$ &
    $0.368 \pm 0.232$\\
    NM3     & $0.391 \pm 0.209$   & 
    $0.171 \pm 0.436$ & 
    $0.622 \pm 0.201$ & 
    $0.295 \pm 0.212$ & 
    $0.447 \pm 0.182$\\
    NUS1     & $0.871 \pm 0.129$   & 
    $\textbf{0.809} \pm 0.428$ & 
    $\textbf{0.969} \pm 0.108$ & 
    $0.895 \pm 0.130$ & 
    $\textbf{0.895} \pm 0.150$\\
    NUS2     & $0.791 \pm 0.182$   & 
    $\textbf{0.736} \pm 0.232$ & 
    $\textbf{0.824} \pm 0.177$ & 
    $0.770 \pm 0.166$ & 
    $\textbf{0.661} \pm 0.208$\\
    CC & $\textbf{0.957} \pm 0.076$   & 
    $0.126 \pm 0.351$ & 
    $0.567 \pm 0.214$ & 
    $\textbf{0.945} \pm 0.097$ & 
    $0.347 \pm 0.178$\\
    NCR     & $0.019 \pm 0.149$   &
    $0.000 \pm 0.000$ & 
    $0.155 \pm 0.345$ & 
    $0.013 \pm 0.123$ & 
    $0.000 \pm 0.000$\\
    TLL     & $0.000 \pm 0.000$   & 
    $0.000 \pm 0.000$ & 
    $0.006 \pm 0.089$ & 
    $0.000 \pm 0.000$ & 
    $0.000 \pm 0.000$\\
    RUS     & $0.528 \pm 0.244$   & 
    $0.206 \pm 0.416$ & 
    $0.605 \pm 0.261$ & 
    $0.518 \pm 0.201$ & 
    $0.472 \pm 0.224$\\
    \bottomrule
  \end{tabular}}
  \label{tab:Balance-gmean}
\end{table}
\FloatBarrier
\begin{table}[!htb]
 \caption{F1 values of Balance dataset using various classifiers}
  \centering
  \resizebox{\linewidth}{!}{
  \begin{tabular}{llllll}
    \toprule
    \multicolumn{6}{c}{Balance Dataset}                   \\
    \midrule
    Method     & GradBoost     & SGD & KNN & RF & LR\\
    \midrule
    ENN & $0.935 \pm 0.021$  & 
    $0.933 \pm 0.216$  & 
    $0.944 \pm 0.014$ &   
    $0.943 \pm 0.012$ & 
    $0.949 \pm 0.004$\\
    AKNN     & $0.936 \pm 0.022$ &
    $0.946 \pm 0.004$  &
    $0.952 \pm 0.014$ & 
    $0.944 \pm 0.010$ &
    $0.946 \pm 0.004$  \\
    NM1     & $0.370 \pm 0.253$    & 
    $0.390 \pm 0.580$ & 
    $0.523 \pm 0.196$ & 
    $0.158 \pm 0.231$ & 
    $0.541 \pm 0.256$\\
    NM2     & $0.796 \pm 0.178$    & 
    $0.372 \pm 0.586$ &
    $0.771 \pm 0.121$ & 
    $0.795 \pm 0.169$ &
    $0.405 \pm 0.228$\\
    NM3     & $0.382 \pm 0.254$   & 
    $0.340 \pm 0.536$ & 
    $0.631 \pm 0.194$ &
    $0.250 \pm 0.264$ & 
    $0.482 \pm 0.214$\\
    NUS1     & $0.890 \pm 0.118$   & 
    $0.888 \pm 0.145$ & 
    $\textbf{0.976} \pm 0.088$ & 
    $0.908 \pm 0.099$ & 
    $0.907 \pm 0.100$\\
    NUS2     & $0.930 \pm 0.060$   & 
    $0.877 \pm 0.220$ & 
    $0.949 \pm 0.033$ & 
    $0.930 \pm 0.045$ & 
    $0.920 \pm 0.041$\\
    CC & $0.951 \pm 0.099$   & 
    $0.308 \pm 0.590$ & 
    $0.515 \pm 0.251$ 
    & $0.946 \pm 0.105$ 
    & $0.331 \pm 0.244$\\
    NCR     & $\textbf{0.953} \pm 0.010$   & 
    $0.957 \pm 0.004$ & 
    $0.955 \pm 0.010$ & 
    $\textbf{0.955} \pm 0.006$ & 
    $0.957 \pm 0.003$\\
    TLL     & $\textbf{0.953} \pm 0.012$   &
    $\textbf{0.959} \pm 0.003$ &
    $0.952 \pm 0.011$ &
    $0.953 \pm 0.009$ & 
    $\textbf{0.959} \pm 0.003$\\
    RUS     & $0.533 \pm 0.196$   &
    $0.427 \pm 0.529$ & 
    $0.566 \pm 0.286$ 
    & $0.515 \pm 0.251$ & 
    $0.455 \pm 0.236$\\
    \bottomrule
  \end{tabular}}
  \label{tab:Balance-f1}
\end{table}

\FloatBarrier

\begin{table}[!htb]
 \caption{AUC values of Pima dataset using various classifiers}
  \centering
  \resizebox{\linewidth}{!}{
  \begin{tabular}{llllll}
    \toprule
    \multicolumn{6}{c}{Pima Dataset}                   \\
    \midrule
    \centering
    Method     & GradBoost     & SGD & KNN & RF & LR\\
    \midrule
    ENN & $0.859 \pm 0.068$  & 
    $0.826 \pm 0.079$  & 
    $0.857 \pm 0.067$ &   
    $0.867 \pm 0.060$ & 
    $0.852 \pm 0.068$\\
    AKNN     & $\textbf{0.880} \pm 0.061$ &
    $0.829 \pm 0.116$  & 
    $\textbf{0.887} \pm 0.059$ & 
    $\textbf{0.887} \pm 0.059$ & 
    $\textbf{0.877} \pm 0.056$  \\
    NM1     & $0.746 \pm 0.066$    & 
    $0.736 \pm 0.136$ & 
    $0.714 \pm 0.057$ & 
    $0.727 \pm 0.081$ & 
    $0.753 \pm 0.054$\\
    NM2     & $0.782 \pm 0.080$    &
    $0.735 \pm 0.108$ &
    $0.763 \pm 0.067$ & 
    $0.777 \pm 0.081$ &
    $0.766 \pm 0.067$\\
    NM3     & $0.659 \pm 0.069$   & 
    $0.638 \pm 0.096$ & 
    $0.605 \pm 0.072$ &
    $0.646 \pm 0.067$ & 
    $0.664 \pm 0.093$\\
    NUS1     & $0.836 \pm 0.059$   &
    $0.805 \pm 0.105$ & 
    $0.817 \pm 0.056$ & 
    $0.845 \pm 0.058$ & 
    $0.823 \pm 0.056$\\
    NUS2     & $0.865 \pm 0.075$   &
    $\textbf{0.852} \pm 0.092$ & 
    $0.846 \pm 0.065$ & 
    $0.870 \pm 0.070$ & 
    $0.854 \pm 0.065$\\
    CC     & $0.680 \pm 0.091$   &
    $0.644 \pm 0.095$ & 
    $0.632 \pm 0.078$ & 
    $0.686 \pm 0.088$ & 
    $0.679 \pm 0.090$\\
    NCR     & $0.870 \pm 0.059$   & 
    $0.806 \pm 0.078$ &
    $0.858 \pm 0.053$ & 
    $0.858 \pm 0.055$ & 
    $0.836 \pm 0.063$\\
    TLL     & $0.754 \pm 0.061$   &
    $0.728 \pm 0.121$ & 
    $0.730 \pm 0.064$ & 
    $0.755 \pm 0.073$ & 
    $0.732 \pm 0.059$\\
    RUS     & $0.724 \pm 0.079$   & 
    $0.705 \pm 0.090$ & 
    $0.700 \pm 0.086$ & 
    $0.731 \pm 0.070$ & 
    $0.727 \pm 0.085$\\
    \bottomrule
  \end{tabular}}
  \label{tab:Pima-auc}
\end{table}
\FloatBarrier
\begin{table}[!htb]
 \caption{G-Mean values of Pima dataset using various classifiers}
  \centering
  \resizebox{\linewidth}{!}{
  \begin{tabular}{llllll}
    \toprule
    \multicolumn{6}{c}{Pima Dataset}                   \\
    \midrule
    \centering
    Method     & GradBoost     & SGD & KNN & RF & LR\\
    \midrule
    ENN & $0.853 \pm 0.061$  & 
    $0.822 \pm 0.107$  & 
    $0.854 \pm 0.069$ &   
    $0.860 \pm 0.061$ & 
    $0.852 \pm 0.058$\\
    AKNN     & $0.874 \pm 0.065$ &
    $\textbf{0.837} \pm 0.144$  & 
    $\textbf{0.883} \pm 0.053$ & 
    $\textbf{0.880} \pm 0.058$ & 
    $\textbf{0.873} \pm 0.046$  \\
    NM1     & $0.739 \pm 0.067$    & 
    $0.702 \pm 0.139$ & 
    $0.705 \pm 0.083$ & 
    $0.725 \pm 0.059$ & 
    $0.752 \pm 0.073$\\
    NM2     & $0.778 \pm 0.084$    &
    $0.711 \pm 0.209$ &
    $0.760 \pm 0.086$ & 
    $0.777 \pm 0.083$ &
    $0.765 \pm 0.082$\\
    NM3     & $0.644 \pm 0.080$   & 
    $0.542 \pm 0.300$ & 
    $0.601 \pm 0.107$ &
    $0.638 \pm 0.086$ & 
    $0.660 \pm 0.079$\\
    NUS1     & $0.841 \pm 0.065$   &
    $0.805 \pm 0.120$ & 
    $0.813 \pm 0.063$ & 
    $0.842 \pm 0.064$ & 
    $0.818 \pm 0.071$\\
    NUS2     & $\textbf{0.886} \pm 0.065$   &
    $0.825 \pm 0.138$ & 
    $0.837 \pm 0.065$ & 
    $0.865 \pm 0.071$ & 
    $0.850 \pm 0.057$\\
    CC     & $0.681 \pm 0.072$   &
    $0.594 \pm 0.256$ & 
    $0.628 \pm 0.067$ & 
    $0.682 \pm 0.072$ & 
    $0.676 \pm 0.085$\\
    NCR     & $0.861 \pm 0.064$   & 
    $0.783 \pm 0.119$ &
    $0.854 \pm 0.061$ & 
    $0.858 \pm 0.062$ & 
    $0.832 \pm 0.066$\\
    TLL     & $0.754 \pm 0.079$   &
    $0.709 \pm 0.172$ & 
    $0.729 \pm 0.803$ & 
    $0.749 \pm 0.081$ & 
    $0.715 \pm 0.078$\\
    RUS     & $0.721 \pm 0.076$   & 
    $0.646 \pm 0.177$ & 
    $0.703 \pm 0.062$ & 
    $0.727 \pm 0.075$ & 
    $0.727 \pm 0.083$\\
    \bottomrule
  \end{tabular}}
  \label{tab:Pima-gmean}
\end{table}
\FloatBarrier
\begin{table}[!htb]
 \caption{F1 values of Pima dataset using various classifiers}
  \centering
  \resizebox{\linewidth}{!}{
  \begin{tabular}{llllll}
    \toprule
    \multicolumn{6}{c}{Pima Dataset}                   \\
    \midrule
    \centering
    Method     & GradBoost     & SGD & KNN & RF & LR\\
    \midrule
    ENN & $0.854 \pm 0.073$  & 
    $0.805 \pm 0.140$  & 
    $0.847 \pm 0.079$ &   
    $0.857 \pm 0.068$ & 
    $0.847 \pm 0.072$\\
    AKNN     & $\textbf{0.880} \pm 0.065$ &
    $0.855 \pm 0.077$  & 
    $\textbf{0.883} \pm 0.073$ & 
    $\textbf{0.890} \pm 0.065$ & 
    $0.874 \pm 0.074$  \\
    NM1     & $0.739 \pm 0.088$    & 
    $0.708 \pm 0.205$ & 
    $0.691 \pm 0.085$ & 
    $0.730 \pm 0.077$ & 
    $0.751 \pm 0.078$\\
    NM2     & $0.770 \pm 0.079$    &
    $0.721 \pm 0.216$ &
    $0.746 \pm 0.090$ & 
    $0.773 \pm 0.075$ &
    $0.760 \pm 0.097$\\
    NM3     & $0.639 \pm 0.078$   & 
    $0.597 \pm 0.322$ & 
    $0.587 \pm 0.095$ &
    $0.643 \pm 0.094$ & 
    $0.654 \pm 0.097$\\
    NUS1     & $0.841 \pm 0.059$   &
    $0.802 \pm 0.158$ & 
    $0.820 \pm 0.059$ & 
    $0.852 \pm 0.062$ & 
    $0.821 \pm 0.066$\\
    NUS2     & $0.841 \pm 0.065$   &
    $\textbf{0.869} \pm 0.119$ & 
    $0.868 \pm 0.071$ & 
    $0.892 \pm 0.055$ & 
    $\textbf{0.880} \pm 0.063$\\
    CC     & $0.678 \pm 0.103$   &
    $0.636 \pm 0.255$ & 
    $0.640 \pm 0.094$ & 
    $0.684 \pm 0.086$ & 
    $0.672 \pm 0.101$\\
    NCR     & $0.862 \pm 0.057$   & 
    $0.804 \pm 0.123$ &
    $0.853 \pm 0.066$ & 
    $0.861 \pm 0.067$ & 
    $0.834 \pm 0.079$\\
    TLL     & $0.687 \pm 0.107$   &
    $0.608 \pm 0.317$ & 
    $0.658 \pm 0.097$ & 
    $0.686 \pm 0.092$ & 
    $0.653 \pm 0.100$\\
    RUS     & $0.732 \pm 0.078$   & 
    $0.665 \pm 0.327$ & 
    $0.702 \pm 0.067$ & 
    $0.727 \pm 0.067$ & 
    $0.728 \pm 0.076$\\
    \bottomrule
  \end{tabular}}
  \label{tab:Pima-f1}
\end{table}
\FloatBarrier
\begin{table}[!htb]
 \caption{AUC values of Satimage dataset using various classifiers}
  \centering
  \resizebox{\linewidth}{!}{
  \begin{tabular}{llllll}
    \toprule
    \multicolumn{6}{c}{Satimage Dataset}                   \\
    \midrule
    \centering
    Method     & GradBoost     & SGD & KNN & RF & LR\\
    \midrule
    ENN & $0.838 \pm 0.032$  & 
    $0.556 \pm 0.149$  & 
    $0.895 \pm 0.032$ &  
    $0.832 \pm 0.036$ &
    $0.516 \pm 0.015$\\
    AKNN     & $0.852 \pm 0.045$ &
    $0.566 \pm 0.171$  & 
    $0.902 \pm 0.031$ & 
    $0.852 \pm 0.045$ &
    $0.516 \pm 0.013$  \\
    NM1     & $0.823 \pm 0.042$    &
    $0.566 \pm 0.111$ &
    $0.790 \pm 0.056$ & 
    $0.834 \pm 0.042$ & 
    $0.705 \pm 0.058$\\
    NM2     & $0.831 \pm 0.052$    & 
    $0.649 \pm 0.192$ &
    $0.832 \pm 0.044$ & 
    $0.851 \pm 0.042$ &
    $0.733 \pm 0.043$\\
    NM3     & $0.692 \pm 0.050$   & 
    $0.524 \pm 0.061$ & 
    $0.688 \pm 0.043$ & 
    $0.715 \pm 0.048$ & 
    $0.572 \pm 0.056$\\
    NUS1    & $\textbf{0.994} \pm 0.009$   & $\textbf{0.993} \pm 0.014$ & 
    $\textbf{0.999} \pm 0.004$ &
    $\textbf{0.997} \pm 0.006$ & 
    $\textbf{0.976} \pm 0.023$\\
    NUS2     & $0.897 \pm 0.036$   &
    $\textbf{0.993} \pm 0.014$ & 
    $\textbf{0.920} \pm 0.027$ &
    $\textbf{0.899} \pm 0.035$ & 
    $\textbf{0.862} \pm 0.038$\\
    CC     & $0.912 \pm 0.035$   &
    $0.545 \pm 0.142$ & 
    $0.863 \pm 0.046$ &
    $0.818 \pm 0.044$ & 
    $0.793 \pm 0.048$\\
    NCR     & $0.820 \pm 0.043$   & 
    $0.545 \pm 0.142$ & 
    $0.885 \pm 0.034$ & 
    $0.818 \pm 0.044$ & 
    $0.515 \pm 0.010$\\
    TLL     & $0.760 \pm 0.044$   & 
    $0.543 \pm 0.149$ & 
    $0.833 \pm 0.039$ &
    $0.765 \pm 0.044$ & 
    $0.511 \pm 0.011$\\
    RUS     & $0.868 \pm 0.033$   & 
    $0.657 \pm 0.151$ &
    $0.879 \pm 0.037$ & 
    $0.880 \pm 0.034$ & 
    $0.709 \pm 0.043$\\
    \bottomrule
  \end{tabular}}
  \label{tab:satimage-auc}
\end{table}
\FloatBarrier

\begin{table}[!htb]
 \caption{G-Mean values of Satimage dataset using various classifiers}
  \centering
  \resizebox{\linewidth}{!}{
  \begin{tabular}{llllll}
    \toprule
    \multicolumn{6}{c}{Satimage Dataset}                   \\
    \midrule
    \centering
    Method     & GradBoost     & SGD & KNN & RF & LR\\
    \midrule
    ENN & $0.825 \pm 0.054$  & 
    $0.286 \pm 0.466$  & 
    $0.888 \pm 0.038$ &  
    $0.818 \pm 0.047$ & 
    $0.000 \pm 0.000$\\
    AKNN     & $0.843 \pm 0.050$ & 
    $0.297 \pm 0.524$  &
    $0.899 \pm 0.043$ & 
    $0.841 \pm 0.042$ & 
    $0.000 \pm 0.000$  \\
    NM1     & $0.805 \pm 0.046$    &
    $0.370 \pm 0.405$ & 
    $0.792 \pm 0.042$ &
    $0.817 \pm 0.044$ &
    $0.724 \pm 0.053$\\
    NM2     & $0.817 \pm 0.042$    & 
    $0.504 \pm 0.524$ &
    $0.826 \pm 0.055$ & 
    $0.839 \pm 0.052$ &
    $0.645 \pm 0.055$\\
    NM3     & $0.696 \pm 0.053$   & 
    $0.243 \pm 0.391$ & 
    $0.688 \pm 0.054$ & 
    $0.719 \pm 0.052$ & 
    $0.596 \pm 0.049$\\
    NUS1    & $\textbf{0.995} \pm 0.008$   & $\textbf{0.993} \pm 0.014$ &
    $\textbf{0.999} \pm 0.004$ & 
    $\textbf{0.998} \pm 0.005$ &
    $\textbf{0.976} \pm 0.022$\\
    NUS2     & $0.893 \pm 0.036$   & 
    $0.874 \pm 0.074$ &
    $0.920 \pm 0.033$ & 
    $0.894 \pm 0.032$ &
    $\textbf{0.862} \pm 0.038$\\
    CC     & $0.909 \pm 0.041$   & 
    $0.545 \pm 0.142$ & 
    $0.858 \pm 0.034$ &
    $0.902 \pm 0.047$ & 
    $0.765 \pm 0.041$\\
    NCR     & $0.801 \pm 0.052$   &
    $0.234 \pm 0.469$ &
    $0.879 \pm 0.034$ &
    $0.797 \pm 0.045$ &
    $0.000 \pm 0.000$\\
    TLL     & $0.731 \pm 0.051$   &
    $0.207 \pm 0.479$ &
    $0.824 \pm 0.039$ &
    $0.738 \pm 0.063$ &
    $0.000 \pm 0.000$\\
    RUS     & $0.867 \pm 0.041$   &
    $0.573 \pm 0.363$ & 
    $0.878 \pm 0.034$ & 
    $0.878 \pm 0.043$ & 
    $0.688 \pm 0.046$\\
    \bottomrule
  \end{tabular}}
  \label{tab:satimage-gmean}
\end{table}

\FloatBarrier
\begin{table}[!htb]
 \caption{F1 values of Satimage dataset using various classifiers}
  \centering
  \resizebox{\linewidth}{!}{
  \begin{tabular}{llllll}
    \toprule
    \multicolumn{6}{c}{Satimage Dataset}                   \\
    \midrule
    \centering
    Method     & GradBoost     & SGD & KNN & RF & LR\\
    \midrule
    ENN & $0.974 \pm 0.006$  & 
    $0.886 \pm 0.273$  & 
    $0.985\pm 0.005$ &   
    $0.976 \pm 0.005$ &
    $0.945 \pm 0.003$\\
    AKNN     & $0.977 \pm 0.006$ & 
    $0.887 \pm 0.203$  & 
    $0.986 \pm 0.005$ &  
    $0.979 \pm 0.006$ & 
    $0.944 \pm 0.003$  \\
    NM1     & $0.821 \pm 0.044$    & 
    $0.471 \pm 0.558$ & 
    $0.792 \pm 0.054$ & 
    $0.833 \pm 0.042$ & 
    $0.715 \pm 0.054$\\
    NM2     & $0.825 \pm 0.042$    &
    $0.645 \pm 0.161$ &
    $0.821 \pm 0.057$ &  
    $0.843 \pm 0.047$ &
    $0.705 \pm 0.05$5\\
    NM3     & $0.703 \pm 0.057$   & 
    $0.443 \pm 0.536$ & 
    $0.687 \pm 0.051$ & 
    $0.721 \pm 0.051$ & 
    $0.573 \pm 0.060$\\
    NUS1     & $\textbf{0.994} \pm 0.009$   & $\textbf{0.993} \pm 0.014$ &
    $\textbf{0.999} \pm 0.004$ & 
    $\textbf{0.997} \pm 0.006$ & 
    $\textbf{0.975} \pm 0.024$\\
    NUS2     & $0.975 \pm 0.008$   & $\textbf{0.965} \pm 0.019$ &
    $0.975 \pm 0.007$ & 
    $0.975 \pm 0.008$ & 
    $\textbf{0.968} \pm 0.008$\\
    CC     & $0.913 \pm 0.030$   & 
    $0.723 \pm 0.115$ &
    $0.863\pm 0.036$ &
    $0.975 \pm 0.006$ & 
    $0.770 \pm 0.053$\\
    NCR     & $0.971 \pm 0.006$   &
    $0.901 \pm 0.224$ & 
    $0.978\pm 0.005$ & 
    $0.975 \pm 0.006$ &
    $0.946 \pm 0.002$\\
    TLL     & $0.966 \pm 0.006$   &
    $0.897 \pm 0.224$ & 
    $0.969 \pm 0.008$ & 
    $0.969 \pm 0.006$ & 
    $0.949 \pm 0.00$2\\
    RUS     & $0.868 \pm 0.033$   &
    $0.635 \pm 0.197$ & 
    $0.871 \pm 0.041$ & 
    $0.877 \pm 0.036$ & 
    $0.672 \pm 0.060$\\
    \bottomrule
  \end{tabular}}
  \label{tab:satimage-f1}
\end{table}
\FloatBarrier

\begin{table}[!htb]
 \caption{AUC values of Ionosphere dataset using various classifiers}
  \centering
  \resizebox{\linewidth}{!}{
  \begin{tabular}{llllll}
    \toprule
    \multicolumn{6}{c}{Ionosphere Dataset}                   \\
    \midrule
    \centering
    Method     & GradBoost     & SGD & KNN & RF & LR\\
    \midrule
    ENN & $0.916 \pm 0.072$  & $0.842 \pm 0.089$  & $0.823 \pm 0.089$ &   $0.929 \pm 0.050$ & $0.839 \pm 0.085$\\
    AKNN     & $0.918 \pm 0.058$ & $0.845 \pm 0.085$  & $0.820 \pm 0.078$ &  $0.925 \pm 0.052$ & $0.837 \pm 0.089$  \\
    NM1     & $0.914 \pm 0.084$    & $0.804 \pm 0.119$ & $0.794 \pm 0.095$ & $0.926 \pm 0.071$ & $0.808 \pm 0.084$\\
    NM2     & $\textbf{0.936} \pm 0.064$    & $0.809 \pm 0.125$ &$0.777 \pm 0.111$ &  $0.940 \pm 0.056$ &$0.823 \pm 0.095$\\
    NM3     & $0.898 \pm 0.079$   & $0.768 \pm 0.200$ & $0.771 \pm 0.096$ & $0.903 \pm 0.082$ & $0.797 \pm 0.099$\\
    NUS1     & $0.935 \pm 0.059$   & $\textbf{0.859} \pm 0.171$ & $\textbf{0.889} \pm 0.075$ & $0.940 \pm 0.060$ & $\textbf{0.856} \pm 0.104$\\
    NUS2     & $0.933 \pm 0.073$   & $\textbf{0.868} \pm 0.208$ & $\textbf{0.879} \pm 0.068$ & $0.940 \pm 0.059$ & $\textbf{0.878} \pm 0.065$\\
    NCR     & $0.934 \pm 0.059$   & $\textbf{0.859} \pm 0.109$ & $0.859 \pm 0.098$ & $\textbf{0.943} \pm 0.063$ & $0.854 \pm 0.084$\\
    TLL     & $0.913 \pm 0.083$   & $0.834 \pm 0.116$ & $0.780 \pm 0.102$ & $0.920 \pm 0.075$ & $0.825 \pm 0.093$\\
    RUS     & $0.904 \pm 0.091$   & $0.831 \pm 0.103$ & $0.818 \pm 0.088$ & $0.910 \pm 0.074$ & $0.828 \pm 0.077$\\
    CC     & $0.902 \pm 0.090$   & $0.812 \pm 0.118$ & $0.791 \pm 0.102$ & $0.908 \pm 0.084$ & $0.818 \pm 0.085$\\
    \bottomrule
  \end{tabular}}
  \label{tab:Ionosphere-auc}
\end{table}
\FloatBarrier
\begin{table}[!htb]
 \caption{G-mean values of Ionosphere dataset using various classifiers}
  \centering
  \resizebox{\linewidth}{!}{
  \begin{tabular}{llllll}
    \toprule
    \multicolumn{6}{c}{Ionosphere Dataset}                   \\
    \midrule
    \centering
    Method     & GradBoost     & SGD & KNN & RF & LR\\
    \midrule
    ENN & $0.916 \pm 0.073$  &
    $0.835 \pm 0.097$  & 
    $0.802 \pm 0.111$ &   
    $0.927 \pm 0.052$ &
    $0.825 \pm 0.103$\\
    AKNN     & $0.912 \pm 0.059$ &
    $0.839 \pm 0.092$  &
    $0.798 \pm 0.099$ &  
    $0.923 \pm 0.054$ & 
    $0.822 \pm 0.105$  \\
    NM1     & $0.916 \pm 0.079$    &
    $0.796 \pm 0.133$ & 
    $0.775 \pm 0.117$ & 
    $0.925 \pm 0.071$ &
    $0.798 \pm 0.096$\\
    NM2     & $0.932 \pm 0.053$    &
    $0.801 \pm 0.132$ &
    $0.753 \pm 0.128$ & 
    $0.939 \pm 0.057$ &
    $0.811 \pm 0.107$\\
    NM3     & $0.901 \pm 0.088$   &
    $0.750 \pm 0.268$ &
    $0.748 \pm 0.117$ &
    $0.902 \pm 0.084$ & 
    $0.789 \pm 0.104$\\
    NUS1     & $0.932 \pm 0.059$   &
    $\textbf{0.859} \pm 0.299$ &
    $\textbf{0.858} \pm 0.083$ &
    $\textbf{0.944} \pm 0.052$ &
    $\textbf{0.853} \pm 0.104$\\
    NUS2     & $0.927 \pm 0.075$   &
    $0.839 \pm 0.385$ &
    $\textbf{0.870} \pm 0.079$ &
    $0.939 \pm 0.061$ & 
    $\textbf{0.870} \pm 0.072$\\
    CC     & $0.896 \pm 0.073$   &
    $0.801 \pm 0.153$ &
    $0.768 \pm 0.130$ & 
    $0.901 \pm 0.082$ & 
    $0.808 \pm 0.099$\\
    NCR     & $\textbf{0.934} \pm 0.048$   &
    $0.853 \pm 0.122$ & 
    $0.846 \pm 0.117$ &
    $0.942 \pm 0.065$ & 
    $0.844 \pm 0.095$\\
    TLL     & $0.913 \pm 0.071$   &
    $0.825 \pm 0.129$ & 
    $0.752 \pm 0.129$ & 
    $0.918 \pm 0.079$ & 
    $0.809 \pm 0.113$\\
    RUS     & $0.897 \pm 0.072$   &
    $0.825 \pm 0.111$ & 
    $0.800 \pm 0.107$ & 
    $0.909 \pm 0.075$ & 
    $0.821 \pm 0.084$\\
    
    \bottomrule
  \end{tabular}}
  \label{tab:Ionosphere-gmean}
\end{table}
\FloatBarrier
\begin{table}[!htb]
 \caption{F1 scores of Ionosphere dataset using various classifiers}
  \centering
  \resizebox{\linewidth}{!}{
  \begin{tabular}{llllll}
    \toprule
    \multicolumn{6}{c}{Ionosphere Dataset}                   \\
    \midrule
    \centering
    Method     & GradBoost     & SGD & KNN & RF & LR\\
    \midrule
    ENN & $0.901 \pm 0.090$  & $0.803 \pm 0.117$  & $0.782 \pm 0.131$ &   $0.917 \pm 0.061$ & $0.803 \pm 0.118$\\
    AKNN     & $0.904 \pm 0.071$ & $0.808 \pm 0.110$  & $0.778 \pm 0.118$ &  $0.913 \pm 0.063$ & $0.801 \pm 0.127$  \\
    NM1     & $0.912 \pm 0.084$    & $0.793 \pm 0.119$ & $0.794 \pm 0.095$ & $0.925 \pm 0.071$ & $0.781 \pm 0.115$\\
    NM2     & $0.933 \pm 0.067$    & $0.799 \pm 0.114$ &$0.723 \pm 0.155$ &  $0.937 \pm 0.058$ &$0.794 \pm 0.125$\\
    NM3     & $0.901 \pm 0.077$   & $0.789 \pm 0.129$ & $0.723 \pm 0.143$ & $0.909 \pm 0.075$ & $0.786 \pm 0.117$\\
    NUS1     & $0.932 \pm 0.068$   & $\textbf{0.879} \pm 0.127$ & $\textbf{0.847} \pm 0.094$ & $\textbf{0.943} \pm 0.055$ & $\textbf{0.841} \pm 0.118$\\
    NUS2     & $\textbf{0.937} \pm 0.068$   & $\textbf{0.883} \pm 0.128$ & $\textbf{0.861} \pm 0.090$ & $\textbf{0.943} \pm 0.053$ & $\textbf{0.861} \pm 0.080$\\
    NCR     & $0.929 \pm 0.066$   & $0.844 \pm 0.119$ & $0.833 \pm 0.135$ & $0.939 \pm 0.070$ & $0.830 \pm 0.110$\\
    TLL     & $0.896 \pm 0.105$   & $0.794 \pm 0.147$ & $0.714 \pm 0.161$ & $0.904 \pm 0.093$ & $0.783 \pm 0.134$\\
    RUS     & $0.901 \pm 0.094$   & $0.821 \pm 0.108$ & $0.780 \pm 0.127$ & $0.909 \pm 0.077$ & $0.807 \pm 0.098$\\
    CC     & $0.897 \pm 0.097$   & $0.805 \pm 0.110$ & $0.741 \pm 0.158$ & $0.904 \pm 0.088$ & $0.792 \pm 0.118$\\
    \bottomrule
  \end{tabular}}
  \label{tab:Ionosphere-f1}
\end{table}

\FloatBarrier
\section{Undersampling on Artificial Dataset}

\begin{figure*}[!htb]
  \centering
 
  \includegraphics[scale=.20]{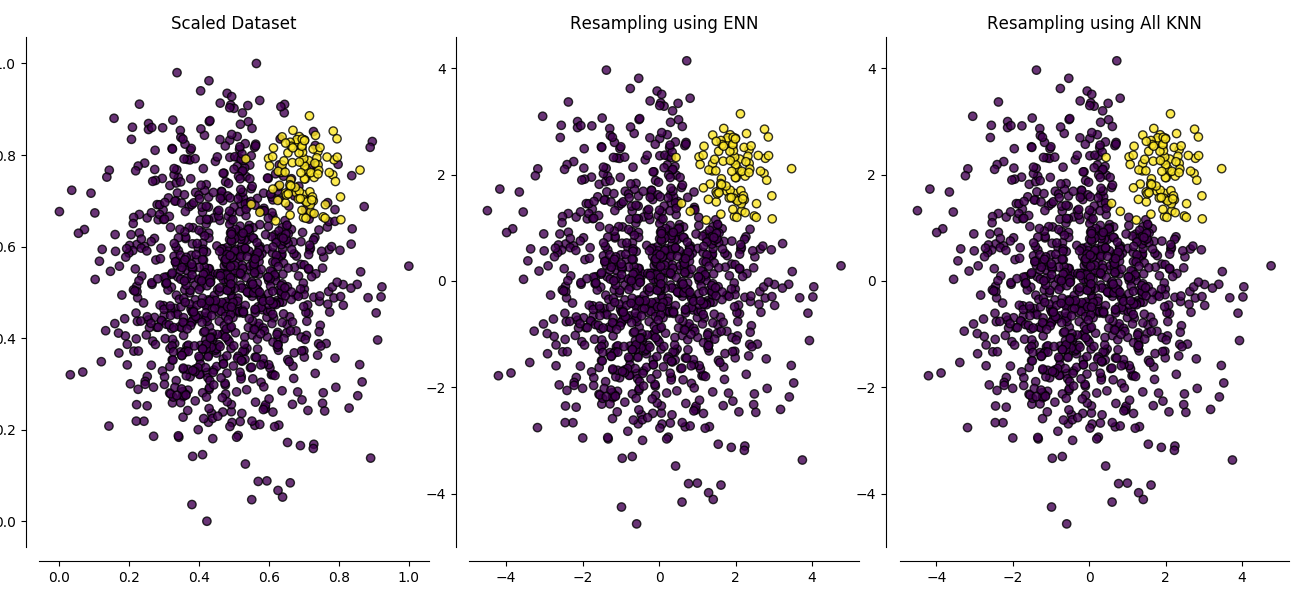}
  \includegraphics[scale=.20]{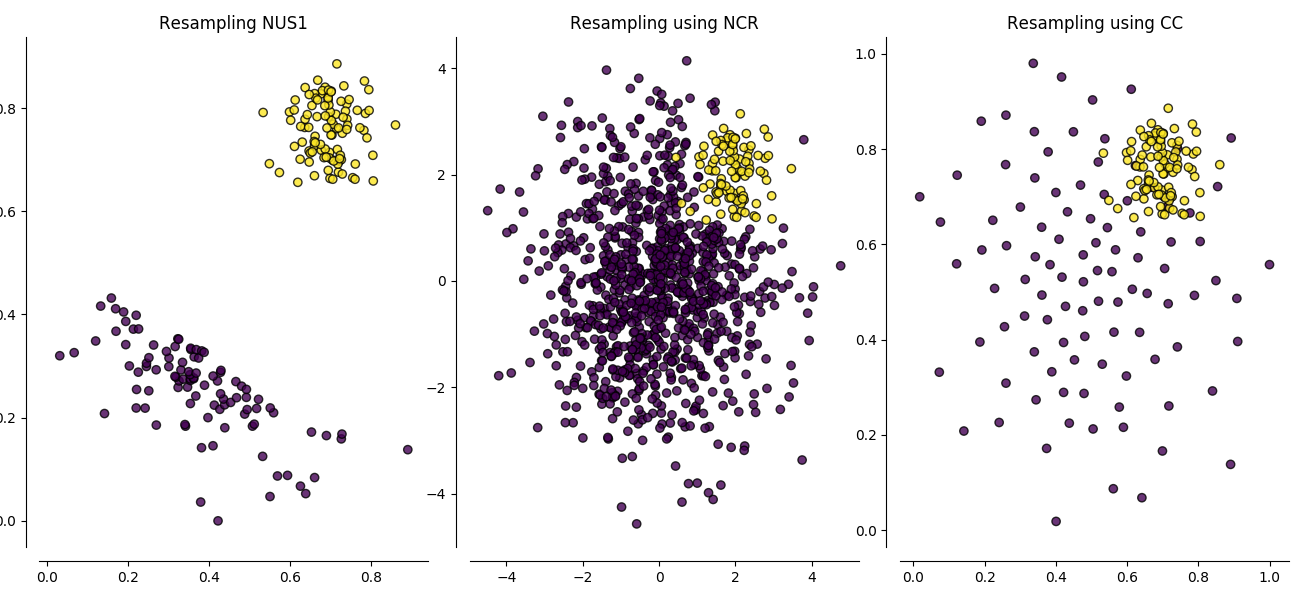}
  \includegraphics[scale=.20]{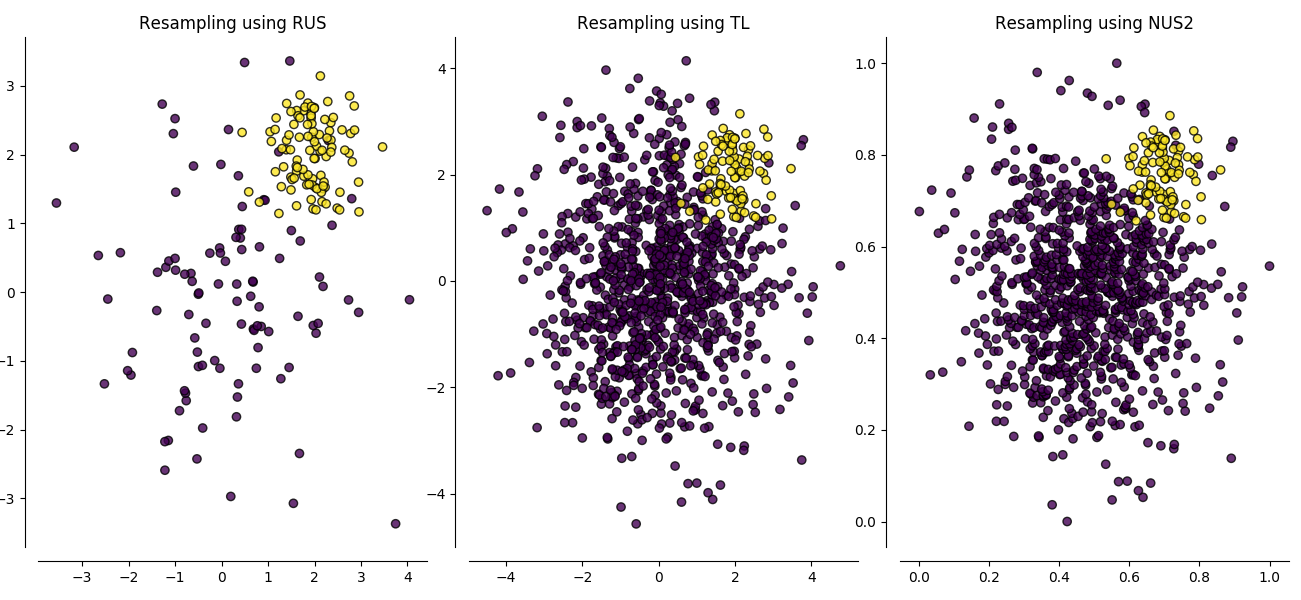}

 \caption{Artificial dataset resampling with various undersamplers}
 \label{fig:fig2}

\end{figure*}

Th datasets on which we experimented so far have lots of features, which makes it difficult to visualize actually how the undersamplers undersample those datasets. Hence we have used two artificial datasets to visualise the effect of different under-samplers using scikit-learn package~\cite{scikit-learn}. The first dataset consists of two features, which makes it easy to plot the dimensions and visualize the data. There are 1000 majority samples and 100 minority samples in the dataset. So, the ratio of majority samples to minority samples is 10:1. The centers of two clusters are [0.0 0.0] and [2.0 2.0] respectively. The standard deviation of the cluster samples from its center are 1.5 and 0.5 each. The effect of each undersampler is shown in the Figure~\ref{fig:fig2}.

\begin{figure*}[!htb]
  \centering

    \includegraphics[scale=.20]{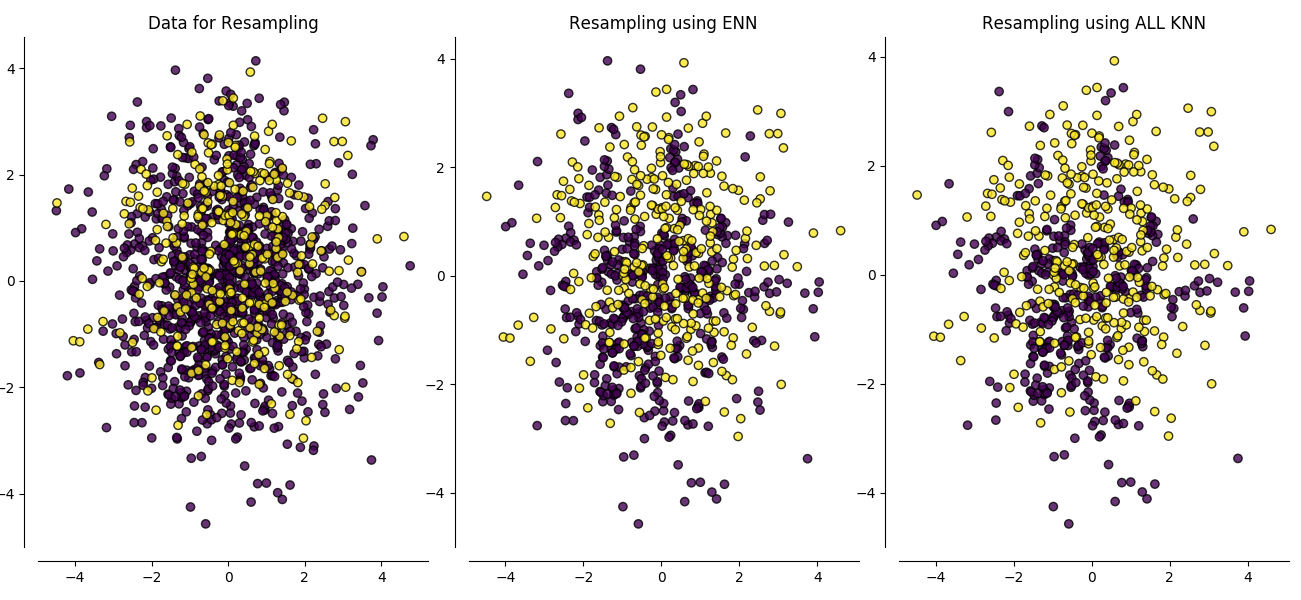}
    \includegraphics[scale=.20]{NUS1_NCR_CC.png}
    \includegraphics[scale=.20]{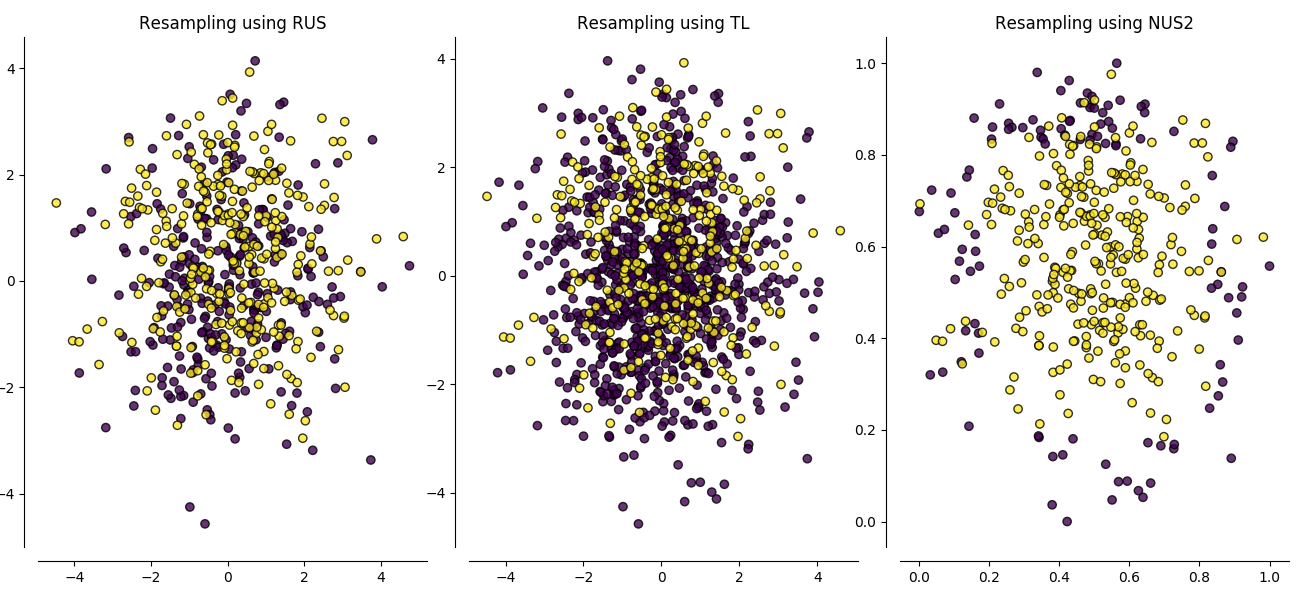}
 \caption{Artificial overlapped dataset resampled by various undersamplers}
 \label{fig:fig3}

\end{figure*}

Next, we generated the second dataset where the majority and minority samples are overlapping in nature.For the second dataset, the ratio of majority to minority is $3.33:1$. In this case, the number of majority samples were same as before but the number of minority samples were $300$. We choose the center of the two classes to be [0.0 0.0] and [0.02 0.05] respectively to introduce the overlapping criteria. The standard deviation of the two cluster samples from the center were $[1.5 1.5]$ respectively. The result of each sampler is shown in Figure~\ref{fig:fig2} and Figure~\ref{fig:fig3}.

\section{Comparison between the proposed algorithms}
It is observed from the classification results and the figures that refered to the effect of each under-sampler in the data that NUS1 performs well when there exists less overlapping in data. NUS1 algorithm actually retains those majority data points which are most distant from most of the minority data points. But in case of overlapped data, there could be some minority samples overlapped with the retained majority data. In case of non-overlapped data, this problem is minimal. Hence, NUS1 makes balanced dataset linearly separable. On the other hand, NUS2 finds the perimeter of minority samples by calculating the average distance of minority samples to its generator samples generated by the model. Then it retains those majority samples that are outside of the perimeter. By this way, overlapping is removed. Hence NUS2 performs better in classifying overlapping data. However, the choosing of distance whether maximum or average is a tunable parameter. We can indirectly verify the nature of the data by these two proposed methods.

\subsection{Case study}
\begin{figure}[!htb]
  \centering

    \includegraphics[scale=.35]{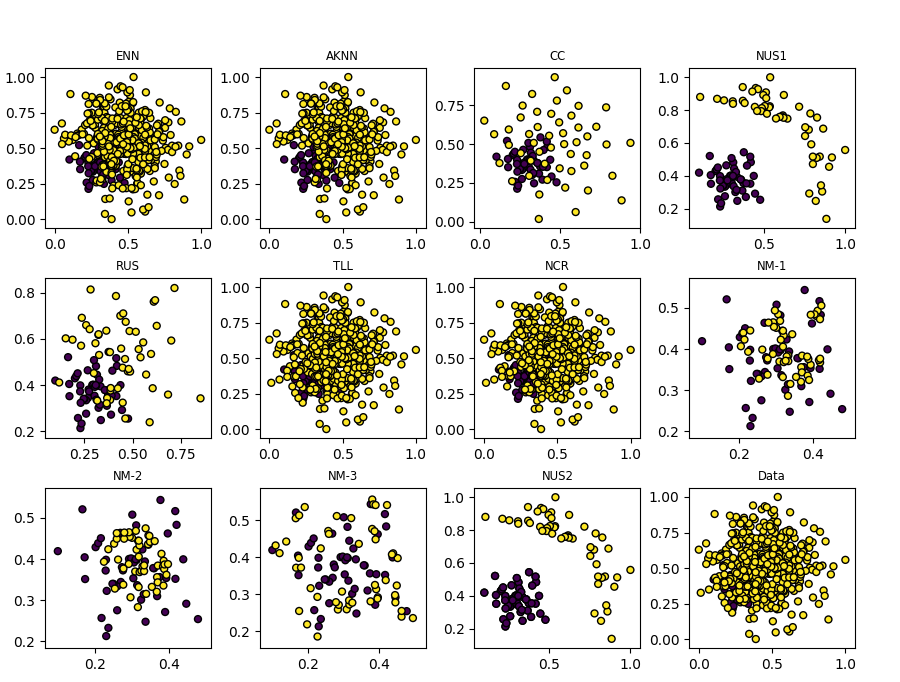}
   
 \caption{NUS-1 and NUS-2 both are trying to choose far samples from minority class. In the figure, the coordinate of the centers are chosen as $[0.0, 0.0]$, $[5.0, 5.0]$ respectively. The standard deviation from center are $2.5$ for the first and $5.5$ for the later one. We have used make\_blob function from scikit-learn~\cite{scikit-learn} to generate the data points.}
 \label{fig:outlier}
\end{figure}

Now we observe a particular case which may arise due to a certain distribution of the data. It may happen that majority class data consists of outliers or data points that are at far distances from the minority data points and also the ratio of majority to minority is very high. In this case, the outliers from the majority data points should be removed first before implementing the proposed hard and soft undersampling algorithms. In Figure ~\ref{fig:outlier} we have generated an artificial dataset using scikit-learn~\cite{scikit-learn} package. The ratio of majority to minority is $500:50$. The two proposed algorithms NUS-1 and NUS-2 always select the 50 points that are located far from minority class at the time of undersampling. In case of outlier, it may happen that the algorithms always choose the outlier data points at the time of undersampling. We have shown the data and effect of undersampling algorithms on data points in Figure ~\ref{fig:outlier}.

\section{Concluding remarks}
In this paper, we proposed two algorithms to solve the class imbalance problem. The main target of this paper is to balance the data i.e. bring down the number of majority samples to the number of minority samples. This approach might result into some drawbacks. If the majority to minority ratio is vary high, there is a high probability of loosing information from majority class. In this scenario, we can use the accuracy of predicting majority samples
as a parameter to choose which batch of majority samples should be considered to mitigate the loss. Future works may address this issue.

\bibliographystyle{unsrt}  
\bibliography{ref}

\begin{thebibliography}{10}

\bibitem{breastcancer}
Bartosz Krawczyk, Mikel Galar, {\L}ukasz Jele{\'n}, and Francisco Herrera.
\newblock Evolutionary undersampling boosting for imbalanced classification of
  breast cancer malignancy.
\newblock {\em Applied Soft Computing}, 38:714--726, 2016.

\bibitem{weather}
Sun Choi, Young~Jin Kim, Simon Briceno, and Dimitri Mavris.
\newblock Prediction of weather-induced airline delays based on machine
  learning algorithms.
\newblock In {\em 2016 IEEE/AIAA 35th Digital Avionics Systems Conference
  (DASC)}, pages 1--6. IEEE, 2016.

\bibitem{Frauddetect}
Wei Wei, Jinjiu Li, Longbing Cao, Yuming Ou, and Jiahang Chen.
\newblock Effective detection of sophisticated online banking fraud on
  extremely imbalanced data.
\newblock {\em World Wide Web}, 16(4):449--475, 2013.

\bibitem{fmeasure}
CJ~Van~Rijsbergen.
\newblock Information retrieval 2nd edition butterworths.
\newblock {\em London available on internet}, 1979.

\bibitem{AUC}
Jin Huang and Charles~X Ling.
\newblock Using auc and accuracy in evaluating learning algorithms.
\newblock {\em IEEE Transactions on knowledge and Data Engineering},
  17(3):299--310, 2005.

\bibitem{g-mean}
Miroslav Kubat, Stan Matwin, et~al.
\newblock Addressing the curse of imbalanced training sets: one-sided
  selection.
\newblock In {\em Icml}, volume~97, pages 179--186. Nashville, USA, 1997.

\bibitem{imb-review}
Yanmin Sun, Andrew~KC Wong, and Mohamed~S Kamel.
\newblock Classification of imbalanced data: A review.
\newblock {\em International Journal of Pattern Recognition and Artificial
  Intelligence}, 23(04):687--719, 2009.

\bibitem{imblearn}
Guillaume Lema{{\^i}}tre, Fernando Nogueira, and Christos~K. Aridas.
\newblock Imbalanced-learn: A python toolbox to tackle the curse of imbalanced
  datasets in machine learning.
\newblock {\em Journal of Machine Learning Research}, 18(17):1--5, 2017.

\bibitem{tomek}
Ivan Tomek.
\newblock A generalization of the k-nn rule.
\newblock {\em IEEE Transactions on Systems, Man, and Cybernetics},
  (2):121--126, 1976.

\bibitem{ncl}
Jorma Laurikkala.
\newblock Improving identification of difficult small classes by balancing
  class distribution.
\newblock In {\em Conference on Artificial Intelligence in Medicine in Europe},
  pages 63--66. Springer, 2001.

\bibitem{smote}
Nitesh~V Chawla, Kevin~W Bowyer, Lawrence~O Hall, and W~Philip Kegelmeyer.
\newblock Smote: synthetic minority over-sampling technique.
\newblock {\em Journal of artificial intelligence research}, 16:321--357, 2002.

\bibitem{smote-1}
Hui Han, Wen-Yuan Wang, and Bing-Huan Mao.
\newblock Borderline-smote: a new over-sampling method in imbalanced data sets
  learning.
\newblock In {\em International conference on intelligent computing}, pages
  878--887. Springer, 2005.

\bibitem{smote-2}
Gustavo~EAPA Batista, Ronaldo~C Prati, and Maria~Carolina Monard.
\newblock A study of the behavior of several methods for balancing machine
  learning training data.
\newblock {\em ACM SIGKDD explorations newsletter}, 6(1):20--29, 2004.

\bibitem{adasyn}
Haibo He, Yang Bai, Edwardo~A Garcia, and Shutao Li.
\newblock Adasyn: Adaptive synthetic sampling approach for imbalanced learning.
\newblock In {\em 2008 IEEE International Joint Conference on Neural Networks
  (IEEE World Congress on Computational Intelligence)}, pages 1322--1328. IEEE,
  2008.

\bibitem{enn}
Dennis~L Wilson.
\newblock Asymptotic properties of nearest neighbor rules using edited data.
\newblock {\em IEEE Transactions on Systems, Man, and Cybernetics},
  (3):408--421, 1972.

\bibitem{nearmiss}
Inderjeet Mani and I~Zhang.
\newblock knn approach to unbalanced data distributions: a case study involving
  information extraction.
\newblock In {\em Proceedings of workshop on learning from imbalanced
  datasets}, volume 126, 2003.

\bibitem{randomforest}
Leo Breiman.
\newblock Random forests.
\newblock {\em Machine learning}, 45(1):5--32, 2001.

\bibitem{gradboost}
Jerome~H Friedman.
\newblock Greedy function approximation: a gradient boosting machine.
\newblock {\em Annals of statistics}, pages 1189--1232, 2001.

\bibitem{knn-1}
RO~Duda and PE~Hart.
\newblock Pattern classification and scene analysis--john wiley \& sons.
\newblock {\em New York, NY}, 1973.

\bibitem{sgd}
Tong Zhang.
\newblock Solving large scale linear prediction problems using stochastic
  gradient descent algorithms.
\newblock In {\em Proceedings of the twenty-first international conference on
  Machine learning}, page 116. ACM, 2004.

\bibitem{lr}
Raymond~E Wright.
\newblock Logistic regression.
\newblock 1995.

\bibitem{scikit-learn}
F.~Pedregosa, G.~Varoquaux, A.~Gramfort, V.~Michel, B.~Thirion, O.~Grisel,
  M.~Blondel, P.~Prettenhofer, R.~Weiss, V.~Dubourg, J.~Vanderplas, A.~Passos,
  D.~Cournapeau, M.~Brucher, M.~Perrot, and E.~Duchesnay.
\newblock Scikit-learn: Machine learning in {P}ython.
\newblock {\em Journal of Machine Learning Research}, 12:2825--2830, 2011.

\bibitem{numpy}
Stefan Van Der~Walt, S~Chris Colbert, and Gael Varoquaux.
\newblock The numpy array: a structure for efficient numerical computation.
\newblock {\em Computing in Science \& Engineering}, 13(2):22, 2011.

\bibitem{matplotlib}
J.~D. Hunter.
\newblock Matplotlib: A 2d graphics environment.
\newblock {\em Computing in Science \& Engineering}, 9(3):90--95, 2007.

\bibitem{keras}
Fran\c{c}ois Chollet et~al.
\newblock Keras.
\newblock \url{https://keras.io}, 2015.

\bibitem{au-roc}
Tom Fawcett.
\newblock An introduction to roc analysis.
\newblock {\em Pattern recognition letters}, 27(8):861--874, 2006.

\bibitem{dataset}
Zejin Ding.
\newblock Diversified ensemble classifiers for highly imbalanced data learning
  and their application in bioinformatics.
\newblock 2011.

\bibitem{Dua:2019}
Dheeru Dua and Casey Graff.
\newblock {UCI} machine learning repository, 2017.

\end{thebibliography}

%

\begin{IEEEbiography}[{\includegraphics[width=0.75in,height=1in,clip,keepaspectratio]{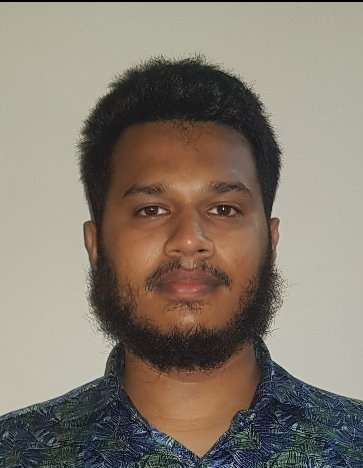}}]{Md. Adnan Arefeen}
Md. Adnan Arefeen is currently working as a lecturer at United Intenational University, Bangladesh. He completed his graduation from Bangldesh University of Engneering and Technology. He is persuing his post graduation degree in the same university.
\end{IEEEbiography}
\vskip 0pt plus -1fil
\begin{IEEEbiography}[{\includegraphics[width=0.75in,height=1in,clip,keepaspectratio]{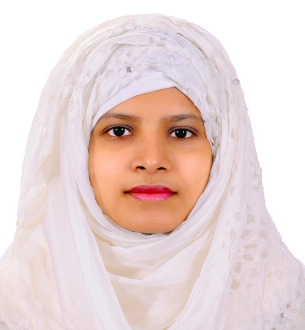}}]{Sumaiya Tabassum Nimi}
Sumaiya Tabassum Nimi completed her Bachelor of Science in Computer Science and Engineering from Bangladesh University of Engineering and Technology, Bangladesh. Currently she is working as a lecturer at United International University.
\end{IEEEbiography}
\vskip 0pt plus -1fil

\begin{IEEEbiography}[{\includegraphics[width=0.75in,height=1in,clip,keepaspectratio]{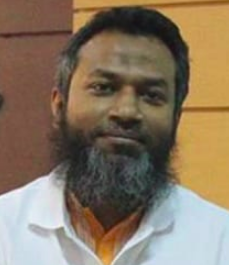}}]{M Sohel Rahman}
Prof. Dr. M Sohel Rahman received his PhD from King's College, University of London. Currently, he is working as a professor at Bangladesh University of Engineering and Technology. His research area includes bioinformatics, algorithms, strings, musicology, graph theory, machine learning.
\end{IEEEbiography}




\end{document}